# Machine Learning for Public Good: Predicting Urban Crime Patterns to Enhance Community Safety

Sia Gupta,[1] Simeon Sayer[2]

[1] *Lynbrook High School, San Jose, CA 95129, USA*

[2] *Faculty of Arts and Sciences, Harvard University, Cambridge, MA 02138, USA*

Ⅰ. **Abstract**

In recent years, urban safety has become a paramount concern for city planners and law enforcement agencies. Accurate prediction of likely crime occurrences can significantly enhance preventive measures and resource allocation. However, many local law enforcement departments lack the tools to analyze and apply advanced AI and ML techniques that can support city planners, neighborhood watch programs, and police and community safety leaders to take proactive steps towards overall community safety.

This paper explores the effectiveness of machine learning techniques to predict spatial and temporal patterns of crimes in large cities and urban areas. Leveraging police dispatch call data from San Jose, CA, the research goal is to achieve a high degree of accuracy in categorizing calls into priority levels particularly for more dangerous situations that require an immediate law enforcement response. This categorization is informed by the time, place, and nature of the call. The research steps include data extraction and preprocessing, feature engineering, exploratory data analysis, implementation, optimization and tuning of different supervised machine learning models and neural networks. The accuracy and precision are examined for different models and features at varying granularity of crime categories and location precision.

The results demonstrate that when compared to a variety of other models, Random Forest classification models are most effective in identifying dangerous situations and their corresponding priority levels with high accuracy (Accuracy = 85%, AUC = 0.92) at a local level while ensuring a minimum amount of false negatives. While further research and data gathering is needed to include other social and economic factors, these results provide valuable insights for law enforcement agencies to optimize resources, develop proactive deployment approaches, and adjust response patterns to enhance overall public safety outcomes in an unbiased way.

## Ⅱ. Introduction

With rising crime in many cities, combined with staffing challenges and limited resource budgets, many law enforcement departments and city governments struggle to improve the safety of their communities. With the advancement of machine learning techniques, it is now possible to analyze vast amounts of data to uncover patterns and make reasonable predictions about future crime patterns [1, 2, 3]. While predicting the exact nature of crime is still science fiction, even broad patterns can provide valuable insights that institutions can use to deploy thoughtful and proactive crime prevention strategies.

Specifically, this study aims to develop and test the effectiveness of several prediction models that can forecast crime patterns, and to identify the models that demonstrate greatest accuracy. Using crime and police call data records from the city of San Jose, CA– one of the largest cities in the country– the goal is to provide law enforcement agencies with tools that can help in resource allocation, preventative planning, and prioritizing real time responses. For example, this research can be developed into a mobile application to deploy for field uses in patrol planning, community safety groups, urban planning, event planning, teen safety, and even property and commercial insurance.

The data used for this study is a mix of numerical and categorical data representing historical records with offense dates, times, nature, and approximate location. After transforming and labeling the data, this study uses supervised learning techniques to classify new reported incidents into severity categories (1 = life-threatening to 6 = low-level offenses). The predictions are tested for accuracy, precision and recall with varying geographical precision. Finally, this paper examines the ethical concerns associated with employing such prediction algorithms for law enforcement, and suggests further areas for further research and additional data.

## Ⅲ. Background

Research on crime prediction using machine learning (ML) and deep learning (DL) encompasses a wide range of approaches and methodologies. These studies provide a robust foundation for understanding the potential and challenges of using ML and DL in crime prediction and form important input into this research study.

Jenga *et al.*'s 2023 paper [1] explores the effectiveness of ML techniques such as Random Forest classification models and artificial neural networks (ANNs) in predicting crime categories across existing publications. These models highlight the need for incorporating additional data types to optimize resource allocation and proactive deployment in law enforcement. Mandalapu *et al.*'s study [2] similarly categorizes traditional methods, such as logistic regression and decision trees, alongside advanced approaches like ANNs and convolutional neural networks (CNNs). It identifies critical gaps, including the necessity for real-time data processing and the integration of socio-economic factors, calling for future research to address these challenges.

There have been many approaches to determining police allocation using machine learning. Downs *et al.* [3] critique traditional resource allocation methods in law enforcement, which often rely on implicit decision-making. They argue for a data-driven approach that considers crime severity and resource requirements to achieve equitable and effective resource distribution. And in their 2021 paper, Shah and Bhagat [5] integrate ML with computer vision techniques to offer a novel approach to crime forecasting. By analyzing CCTV footage alongside historical crime data, this research predicts crime hotspots, illustrating the potential of combining visual data with traditional records for real-time surveillance and urban safety planning.

Some work attempts to use a similar data source to this study. For instance, Chohlas-Wood *et al.'s* analysis in [4] of 911 call data from New York City uses ML to identify temporal patterns and forecast future call volumes, demonstrating how predictive models can enhance emergency response efficiency by anticipating peak call times and optimizing resource allocation. However, Chohlas-Wood relies on the assumption that high call count is correlated with high priority. This paper's analysis of San Jose data is predicting not the number, but the importance of these calls, which is arguably more important in determining resource allocation.

This paper also attempts to account for underlying bias within this data. The research by Rotaru, Huang and Li *et al.* [6] focuses on the biases in law enforcement data when predicting urban crime, revealing inherent enforcement biases along racial and socio-economic lines. It underscores the ethical challenges of using biased data in predictive models and the necessity of addressing these biases to ensure fair policing practices. Similarly the 2018 paper by Thirumalaiswamy, Vliet and Sun from the University of Pennsylvania [7] emphasizes the influence of socio-economic and demographic factors while developing models to predict crime in Philadelphia. In particular it highlights the importance of localized models that account for city-specific characteristics to achieve accurate predictions. And Saeed and Abdulmohsin's 2023 study [8] demonstrates the role of text data mining in crime prediction by analyzing textual descriptions from police reports and social media to predict crime trends and hotspots, highlighting the potential of unstructured text data.

Predicting calls is not the same thing as predicting the crimes themselves– although there is existing work in this area. Matt Woods' review article [9] for the University of Chicago showcases the capability of algorithms to predict crimes up to a week in advance while revealing biases in police responses. This research calls for developing algorithms that mitigate these biases and promote fairness, emphasizing transparency and accountability in predictive models. Dakalbab et al. 's systematic literature review [10] covers statistical models, ML algorithms, and neural networks and identifies trends, challenges, and future directions, emphasizing interdisciplinary research and diverse data integration. Lastly, in Walczak's 2021 research article

[11], the authors explore the use of neural networks in crime prediction and police decision-making, highlighting their ability to handle complex data relationships and the potential for combining them with other ML techniques for improved accuracy.

These works were important in the contextualizing existing approaches and identifying models of interest. However, this study, while building off these important contributions, particularly in regard to bias and fairness, attempts to find new predictive properties of unexplored data features.

## Ⅳ. Dataset

**4.1 Data Collection**

The data used for this research was the police dispatch data from the city of San Jose [12], CA, the 12th largest city in the country with a population of 932,636 residents. The San Jose Police Department (SJPD) receives, according to our data, between 0.3 million to 0.4 million inbound calls annually that require a law enforcement response. SJPD publishes this data on their public database. The patterns are representative of a large US metro, and many major cities like San Francisco, New York, Philadelphia and Chicago also publish similar data (with some variations in frequency and data formats).

The San Jose data is available from 2013 to 2023. For this paper, the most recent three full years from 2021 to 2023 were used to account for the fact that crime patterns change over time and recent data is most relevant. They are downloaded as CSV files and were combined into a common pandas dataframe using Python. The resulting data frame contains over 900,000 samples for training. Each record represents an incident and includes 15+ fields such as the timestamp of the offense, report filing date, call number, call ID, call priority, call reason descriptions and codes, and address range or street intersection. The raw data is a mix of integers, floats, alphanumeric strings and objects, so several pre-processing steps were required to prepare the data for analysis and training.

**4.2 Data Pre-processing and Cleanup**

First the data was de-duplicated based on unique call IDs, and records with null or missing values were removed to ensure integrity and to prevent shape errors. The street address ranges and intersections for each call were transformed into geo-spatial coordinates using a third party geocoding service and stored as LATITUDE and LONGITUDE within floats.

**4.3 Data analysis and Visualization**

Initial analysis involved visualizing and examining the distribution of calls by priority levels (1-6), weekdays, time of day, and months to identify any underlying patterns and anomalies. Category 1 calls (life threatening situations like firearms discharge, violent assault) were less frequent, as expected since these are rare. The highest frequency of calls were in category 2 (dangerous situations like silent alarm or assault with deadly weapon) and category 3 (alarming situations like gang activity or threatening person). The highest frequency of category 2 and category 3 categories is reflective of the self-selection made by citizens calling in for 911 help in such scenarios, and less often for category 4, 5, and 6 categories like wellness checks, abandoned cars, and others.

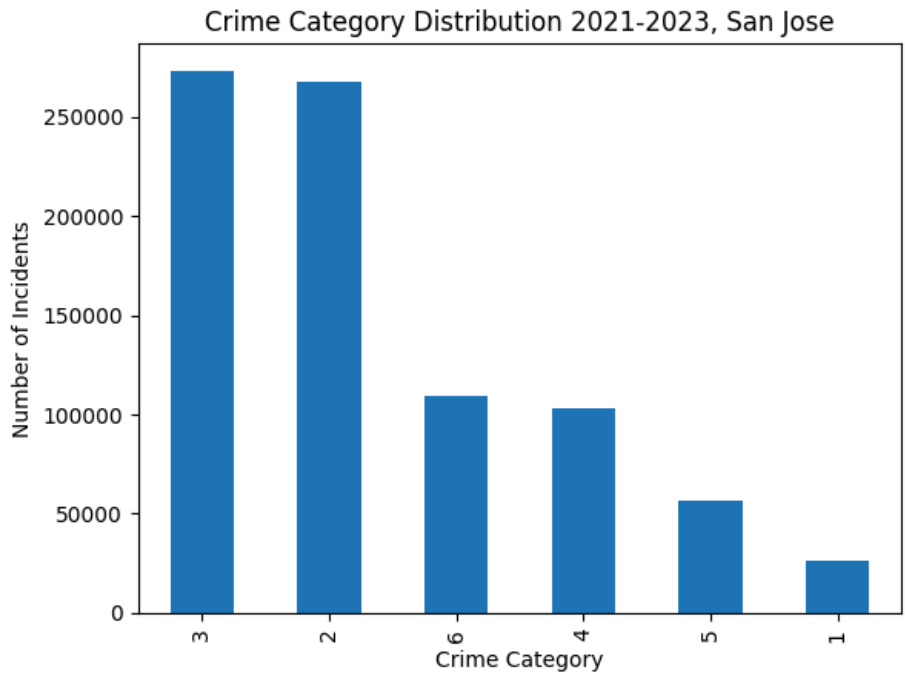

**Fig.1** Crime incident distribution by category one through six, with the most severe incidents labeled as one and the least severe incidents labeled as six.

### 4.4 Feature Engineering, Extraction and Scaling

Prior to model training, some further ML model specific data steps were conducted. The date, time and year were extracted from the timestamp and transformed into day of the week labels (WEEKDAY_NUM) and 30-minute time windows (TIME_CHUNK). To create a citywide grid of location points, the latitude and longitude coordinates were rounded to 1 and 2 decimal places, which corresponds approximately to a 2x2 square mile unit and a 1x1 square mile unit, and stored as additional columns. This simultaneously ensures a consistent size of the region, preventing overfitting, as well as ignoring human centric labels which are prone to bias, such as ZIP codes and neighborhoods. Finally, the key input data features used for training TIME_CHUNK, OFFENSE_WEEKDAY, CALL_TYPE, PRIORITY, LATITUDE, LONGITUDE were retained in the dataframe and then scaled using Python sklearn's preprocessing StandardScaler.

```
#define input features, without location coordinates for citywide prediction
features = df[['OFFENSE_WEEKDAY','TIMECHUNK_NUM','CALLTYPE_NUM']]

#define target as all 6 crime categories
target = df['PRIORITY']

#split the data. Use _m as indicator for multiclass classification
X_train_m, X_test_m, y_train_m, y_test_m = train_test_split(features,target,test_size=0.2,random_state=42)
```

```python
#scale the input features
from sklearn.preprocessing import StandardScaler
scaler = StandardScaler()
scaler.fit(X_train_m)
X_train_m = scaler.transform(X_train_m)
X_test_m = scaler.transform(X_test_m)
```

**Fig.2** Python implementation of feature engineering, extraction, and split of data into training and testing datasets.

In practice, the urgency of a law enforcement response and officer dispatch is usually a binary decision. To model this, an additional binary class (DANGEROUS_SITUATIONS) was derived with the 'TRUE' label being reflected by the highest danger categories, 1 and 2, and 'FALSE' reflecting less serious crime categories 3, 4, 5, and 6. The decision to do this was made after experimentation between a multiple classification and a regression model on the category values. The binary model, as detailed in the following sections, performed strongest.

Finally, the dataset was split into training (80%) and testing sets (20%) using a random train_test split. This data was trained and the class weights derived on this random split, to remove sample bias from the training data used for the prediction models.

## V. Crime Prediction Methodology and Models

### 5.1 Exploring Linear and Random Forest Regressions

The first step of the research was to explore basic linear and random forest regression models and assess their effectiveness in predicting the crime category, compared to a classification approach. Ultimately, both regression and classification must be tested in order to determine the most effective approach to building a predictive model for this data.

```python
#citywide regression approach for multiclass classification
from sklearn.linear_model import LinearRegression
from sklearn.metrics import mean_squared_error, r2_score
Linearmodel_citywide = LinearRegression()
Linearmodel_citywide.fit(X_train_m,y_train_m)
y_pred_linear = Linearmodel_citywide.predict(X_test_m)
RFregressionmodel_citywide = RandomForestRegressor()
RFregressionmodel_citywide.fit((X_train_m,y_train_m)
y_pred_rf = Linearmodel_citywide.predict(X_test_m)
```

**Fig.3** Python implementation of linear regression model training.

The linear regression model did not provide great performance (r-squared value of 0.02 and mean-squared error of 1.91) while Random Forest Regression provided a good starting point

(r-square of 0.81 and MSE of 0.37). To further improve these results, the remainder of the research was focused on testing different classification techniques, including Random Forest.

## 5.2 Multiclass Classification for Citywide Crime Category Prediction

The next step was to set up a multiclass classification model to predict the granular (1-6) levels of crime categories. This step was also crucial in establishing the imbalances in the data and feasibility of predicting such granular crime categories. For this phase, the predictions were conducted at a citywide level (*i.e.* not taking location features into consideration).

After exploring literature for models such as logistic regression, decision trees, random forests, and neural networks, a Random Forest Classifier was chosen as a primary approach for multi-class classification at citywide level. The Random Forest algorithm is an ensemble learning method that operates by constructing multiple decision trees during training and outputting the mode of the classes for classification tasks. Random Forest builds several decision trees and merges them to get a more accurate and stable prediction. Each tree is trained on a random subset of the data and features.

```python
from sklearn.ensemble import RandomForestClassifier
RFmodel_citywide = RandomForestClassifier(n_estimators=100,random_state=42,class_weight='balanced')
RFmodel_citywide.fit(X_train,y_train)
y_pred = RFmodel_citywide.predict(X_test)
```

**Fig.4** Python implementation of Random Forest classifier for multi-class classification.

## 5.3 Using Binary Classification for Modeling Dangerous Situations

Recognizing the need to enhance the performance and practical use for a law enforcement response, the six crime categories were converted into a binary class, delineating all calls into high-priority (dangerous) and low-priority (less dangerous) categories (DANGEROUS_SITUATIONS = 1 or 0). This transformation allowed for a more focused analysis on Binary Classification of dangerous crime occurrences, which often require priority response.

```python
df['DANGEROUS_SITUATION'] = df['PRIORITY'].apply(lambda x: 1 if x == 1 or x == 2 else 0)
```

**Fig.5** Python implementation of dangerous and non-dangerous situation mapping.

Further investigations were conducted using a similar approach of splitting the data into training and testing datasets and implementing logistic regression, decision trees, and random forests for binary classification without incorporating geographical features to simulate citywide predictions. This approach aimed to understand the impact of temporal and categorical data alone on prediction accuracy. Among these, the random forest model stood out due to its robustness against overfitting and its ability to handle unbalanced data, making it an ideal candidate for further refinement. The Random Forest algorithm was optimized for performance using RandomSearchCV to find the appropriate hyperparameters.

```python
from sklearn.model_selection import RandomizedSearchCV
grid_param = {
    'n_estimators': [100, 200, 300],
    'criterion': ['gini', 'entropy'],
    'max_depth': [10,20,30],
    'min_samples_split': [2,3,4],
    'min_samples_leaf': [1,2,3]
}
random_search = RandomizedSearchCV(estimator=RFmodel_citywide,param_distributions=grid_param,cv=5,n_jobs=-1)
random_search.fit(X_train, y_train)
print(random_search.best_params_)
```

**Fig.6** Python implementation of Random Forest algorithm optimized using RandomSearchCV.

**5.4 Incorporating Geographical Features for neighborhood level predictions**

Recognizing the importance of spatial data, latitude and longitude features were incorporated into the prediction approach. Building on the success of the random forest model for predicting DANGEROUS_SITUATION class at a citywide level, the research next incorporated LATITUDE and LONGITUDE as input features into the Random Forest classifier. Different levels of precision (1x1 square mile or 2x2 square miles) for latitude and longitude data were experimented with to measure their impact on model accuracy and determine the optimal granularity.

```python
df['LAT_ROUND1'] = df['LATITUDE'].round(1)
df['LONG_ROUND1'] = df['LONGITUDE'].round(1)
df['LAT_ROUND2'] = df['LATITUDE'].round(2)
df['LONG_ROUND2'] = df['LONGITUDE'].round(2)
features = df[['OFFENSE_WEEKDAY','TIMECHUNK_NUM','CALLTYPE_NUM','LATITUDE','LONGITUDE']]
target = df['DANGEROUS_SITUATION']
X_train, X_test, y_train, y_test = train_test_split(features,target,test_size=0.2,random_state=42)
RFmodel_actual_latlong = RandomForestClassifier(n_estimators=100,random_state=42,class_weight='balanced')
RFmodel_actual_latlong.fit(X_train,y_train)
y_pred = RFmodel_actual_latlong.predict(X_test)
```

**Fig.7** Python implementation of rounding the LATITUDE and LONGITUDE input features and testing the effect of geographical input data on prediction outputs.

## 5.5 Deep Learning Approach

In the final phase of the study, the study explored the capabilities of deep learning by implementing an Artificial Neural Network (ANN). An Artificial Neural Network is a Deep Learning model inspired by the way biological neural networks in the human brain incrementally process information to make decisions. ANNs consist of layers of interconnected nodes, or neurons, that can learn to recognize hidden patterns. ANNs learn through a process called backpropagation, where the model adjusts its weights based on the error of its predictions during training.

Neural networks like ANN require a special model setup. Using sklearn and numpy, the dataset was split into training, validation, and testing sets. To handle class imbalance in the training data, class weights were computed and used during training ensuring that the model did not bias towards the class majority.

Two versions of ANN were created and trained– one for multi-class category identification and one for binary predictions. Both were designed to leverage the combined strength of temporal, categorical, and spatial features to predict crime patterns. The ANN was compiled with the Adam optimizer and binary cross-entropy loss function. The input layer receives the features, including OFFENSE_WEEKDAY, TIMECHUNK_NUM, CALLTYPE_NUM, LATITUDE and LONGITUDE. The network consists of multiple hidden layers with a specified number of neurons. The model that we found to be most optimal for training and accuracy was two hidden layers, each of 15 neurons. Each layer uses the ReLU activation function to introduce non-linearity. And for the output layer: The output layer uses the Sigmoid activation function to predict the probability of a dangerous situation. The model was trained for 25 and 50 epochs with batch size adjustments and went through rigorous hyperparameter tuning to optimize its performance. The model's performance was evaluated on the validation and test sets using accuracy, precision, recall, and other relevant metrics.

```python
from keras.models import Sequential
from keras.layers import Dense
from keras.optimizers import Adam
from keras.metrics import Precision, Recall, AUC
from sklearn.utils import compute_class_weight
import numpy as np
from sklearn.model_selection import train_test_split

#Define features and target
features = df[['OFFENSE_WEEKDAY','TIMECHUNK_NUM','CALLTYPE_NUM','LATITUDE','LONGITUDE']]
target = df['DANGEROUS_SITUATION']
```

```python
#Split the data
X_train, X_temp, y_train, y_temp = 
train_test_split(features,target,test_size=0.2,random_state=42)
X_validate, X_test, y_validate, y_test = 
train_test_split(X_temp,y_temp,test_size=0.5,random_state=42)
#Convert to numpy arrays
X_train = np.array(X_train)
X_validate = np.array(X_validate)
X_test = np.array(X_test)
#Define the ANN model
ann_model_binary = Sequential()
#Add input layer to model
ann_model_binary.add(Dense(units=15,activation='relu',input_shape=(X_train.shap
e[1],)))
#Add 2 hidden layers
ann_model_binary.add(Dense(units=15,activation='relu'))
ann_model_binary.add(Dense(units=15,activation='relu'))
#Add output layer
ann_model_binary.add(Dense(1,activation='sigmoid'))
#Compile model
ann_model_binary.compile(optimizer=Adam(learning_rate=0.004),loss='binary_cross
entropy',metrics=['accuracy',Precision(),Recall()])
#adjust class weights
class_weights = compute_class_weight('balanced',classes = np.unique(y_train),y
= y_train)
class_weights = dict(enumerate(class_weights))
#Fit the model
ann_model_binary.fit(X_train,y_train,epochs=50,validation_data=(X_validate,y_va
lidate),batch_size=64,class_weight=class_weights)
```

**Fig.8** Building an ANN model using Python with two hidden layers (ReLU). Adjusting class weights and fitting the model on the training dataset.

## VI. Results and Discussions

The performance of models was assessed using the test set data using several evaluation metrics:

- *Accuracy:* The proportion of correct predictions out of the total predictions made.
- *Precision and Recall*: Precision measures the accuracy of positive predictions, while Recall measures the ability to identify all positive events.
- *ROC-AUC Score*: The Area Under the Receiver Operating Characteristic Curve provides a measure of the model's performance across all classification thresholds.
- *Confusion Matrix:* A table used to describe the performance of a classification model by showing the true data versus the predicted classification of the model.
- *ROC Curves*: The ROC curves were plotted to visualize the models' performance.

## 6.1 Citywide Multi-Class Classification of all crime categories

The initial attempt to classify crimes into six priority levels at a citywide level yielded a moderate 73% accuracy. Confusion matrices and classification reports indicated challenges in distinguishing between adjacent priority levels except the lowest priority category (category six).

| Category | Precision | Recall | f1-score | Support |
| --- | --- | --- | --- | --- |
| 1 | 0.39 | 0.71 | 0.50 | 5360 |
| 2 | 0.73 | 0.79 | 0.76 | 53777 |
| 3 | 0.82 | 0.63 | 0.71 | 54338 |
| 4 | 0.73 | 0.67 | 0.70 | 20635 |
| 5 | 0.40 | 0.59 | 0.48 | 11444 |
| 6 | 0.98 | 0.99 | 0.99 | 21859 |
| | | Accuracy | 0.73 | 167413 |

**Fig.9** Classification of crime data split across the six severity categories.

## 6.2. Citywide Binary Classification Techniques (without Geolocation features)

Simplifying the problem to a binary classification significantly improved model performance. Metrics such as accuracy, precision, recall, and F1-score displayed effective predictive capabilities. Among the logistic regression, decision tree, and random forest models, the random forest displayed the most effective performance. This model's ability to handle non-linear relationships and interactions between features contributed to its performance.

| | Precision | Recall | f1-score | Support |
| --- | --- | --- | --- | --- |
| 0 | 0.70 | 0.75 | 0.72 | 108276 |
| 1 | 0.47 | 0.41 | 0.44 | 59137 |
| | | Accuracy | 0.63 | 167413 |

**Fig.10** Precision, Recall, and f1-score using logistic regression.

|   | Precision | Recall | f1-score | Support |
|---|---|---|---|---|
| **0** | 0.67 | 0.93 | 0.78 | 108276 |
| **1** | 0.59 | 0.18 | 0.27 | 59137 |
|   | **Accuracy** | | 0.67 | 167413 |

**Fig.11** Precision, Recall, and f1-score using Naive Bayesian.

|   | Precision | Recall | f1-score | Support |
|---|---|---|---|---|
| **0** | 0.95 | 0.82 | 0.88 | 108276 |
| **1** | 0.74 | 0.91 | 0.82 | 59137 |
|   | **Accuracy** | | 0.85 | 167413 |

**Fig.12** Precision, Recall, and f1-score using Decision Trees.

|   | Precision | Recall | f1-score | Support |
|---|---|---|---|---|
| **0** | 0.94 | 0.82 | 0.88 | 108276 |
| **1** | 0.73 | 0.91 | 0.81 | 59137 |
|   | **Accuracy** | | 0.85 | 167413 |

**Fig.13** Precision, Recall, and f1-score using Random Forest.

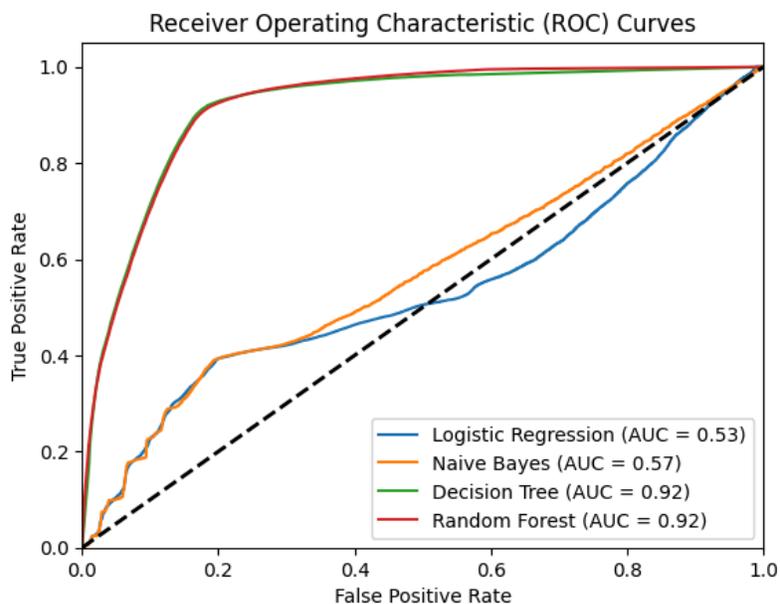

**Fig.14** Receiver Operating Characteristic (ROC) curves for each tested Binary Classification model.

### 6.3 Incorporating Location Features

Adding latitude and longitude features to the Random Forest model further refined predictions. The random forest models were retrained with these feature sets, and the performance was measured using precision, recall and AUC score. Various levels of precision for these coordinates were tested, with narrower granularity, leading to a better prediction of crime hotspots. The research hypothesis stated that including location data with varying precision would yield different levels of accuracy, which is shown in the results.

|   | Precision | Recall | f1-score | Support |
|---|---|---|---|---|
| 0 | 0.88 | 0.86 | 0.87 | 108276 |
| 1 | 0.75 | 0.78 | 0.77 | 59137 |
|   | **Accuracy** | | 0.83 | 167413 |

**Fig.15** Precision, Recall, and f1-score using Random Forest (1x1 square mile grid precision).

|   | Precision | Recall | f1-score | Support |
|---|---|---|---|---|
| 0 | 0.93 | 0.82 | 0.87 | 108276 |
| 1 | 0.73 | 0.88 | 0.80 | 59137 |
|   | **Accuracy** | | 0.84 | 167413 |

**Fig.16** Precision, Recall, and f1-score using Random Forest (2x2 square mile grid precision).

|   | Precision | Recall | f1-score | Support |
|---|---|---|---|---|
| 0 | 0.92 | 0.84 | 0.88 | 108276 |
| 1 | 0.75 | 0.87 | 0.81 | 59137 |
|   | **Accuracy** | | 0.85 | 167413 |

**Fig.17** Precision, Recall, and f1-score using Random Forest (Actual Latitude and Longitude).

### 6.3.4. Deep Learning Model:

Overall, For Binary Classification using ANN, the deep learning approach utilizing a neural network architecture achieved a moderate 62% accuracy and AUC = 0.68 after 50 epochs. However, the multi-class classification yielded better results with 69% accuracy overall, and high AUC scores from 0.82 to 0.84 for the higher priority crime categories (Categories 1-3), and 0.91 to nearly 1.0 for the lower priority categories. The hyperparameters were optimized for both ANNs which yielded very minimal improvements despite significantly increased training time and computational resources.

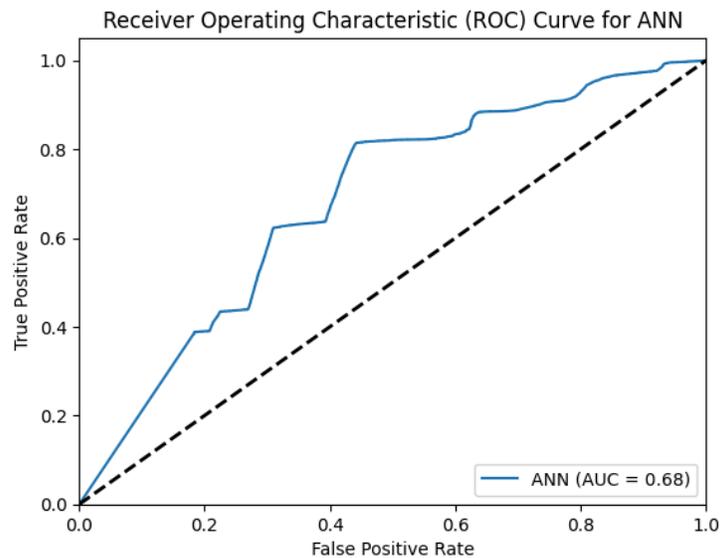

**Fig.18** Binary Classification ROC curve using an ANN.

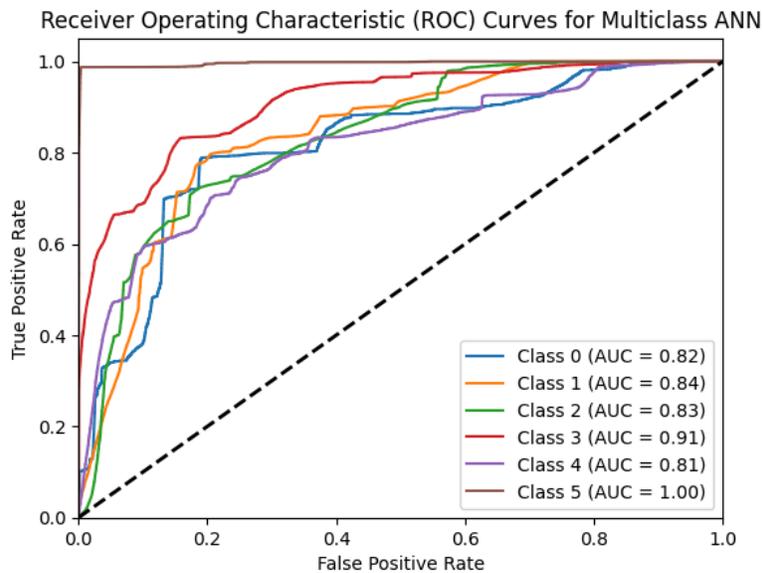

**Fig.19** Multiclass Classification ROC curve using an ANN.

| Category | Precision | Recall | f1-score | Support |
|---|---|---|---|---|
| 1 | 0.57 | 0.03 | 0.05 | 2644 |
| 2 | 0.61 | 0.81 | 0.70 | 26908 |
| 3 | 0.66 | 0.70 | 0.68 | 27177 |
| 4 | 0.76 | 0.56 | 0.65 | 10335 |
| 5 | 0.00 | 0.00 | 0.00 | 5713 |
| 6 | 0.96 | 0.99 | 0.97 | 10930 |
| | | Accuracy | 0.69 | 83707 |

**Fig.20** Crime data split across the six severity categories using an ANN.

## VII. Conclusion

Throughout the study, each model's performance was evaluated using a variety of metrics, including accuracy, precision, recall, and the area under the ROC curve. These metrics provided insights into each model's effectiveness and limitations in different scenarios. The progression from basic classifiers to an advanced deep learning framework showcased the potential of machine learning in enhancing predictive policing strategies. By systematically increasing the complexity and scope of the models, the study offers systematic exploration of all model types, meaning that conclusions from this data translates into actionable insights for law enforcement agencies to optimize resource allocation and proactive policing efforts.

A comparative analysis of the binary classification models revealed that an optimized Random Forest model provides the best results at a citywide level. (AUC = 0.92, Accuracy = 85%). This was higher than the results of the deep learning ANN model (AUC = 0.68, Accuracy = 62%), for the configurations we tried. Additionally, neural networks are black box models, with a low amount of interpretability, meaning that its results, even if higher, would be harder to justify as there would not be the ability to examine which features it was using, and how, prompting questions of bias and fairness. When the dangerous class (1) is considered as the positive class, the RandomForest algorithm demonstrates 0.91 recall. A high recall value is preferred as it minimizes false negatives, because a false negative implies a dangerous situation that was incorrectly classified as non-dangerous. On the other hand, a moderate false positive score implies a suboptimal use of resources (*e.g.* dispatching an officer as priority even if it wasn't a highly dangerous situation).

The research also found that the precision of latitude and longitude data showed a marginally higher accuracy (83%-84%) by reducing the precision of the geographic inputs. In the urban field, higher precision location levels enable faster response times. Given this tradeoff between the precision of the location versus the accuracy of the model, further research is needed to find the optimal tradeoff to enhance the model's utility for law enforcement agencies. This includes the possibility of tracing an exact location for the response team, but only giving a reduced accuracy version to the model.

Further research should also consider testing additional national, state, and local data into the models. This could include local social and demographic data (*e.g.* unemployment, local businesses, income, education, housing shortages), statewide factors *(e.g.* criminal prosecution laws, bail patterns, drug trafficking laws, gun reform, law enforcement budgets), or even national macroeconomic data (*e.g.* recessions, national political climate). Social media also employs the use of consumer data that could be implemented in such a model. Finally, the use of large language models like GPT, Llama or Gemini that use new transformer- based architectures to detect patterns in data, accounting for the dynamic nature of crime, should be explored, especially with new advances in emotion and urgency detection within cutting edge voice models.

This paper presents a case for the integration of advanced technology, specifically machine learning, in public safety strategies. The findings demonstrate that machine learning models can accurately predict crime hotspots, enabling law enforcement to allocate resources more effectively and proactively to address potential crime. As these results are considered, it becomes clear that embracing such technology is not just an option but a necessity for modern policing. The implications of this research call for immediate action: policymakers, community leaders, and law enforcement agencies must collaborate to further research and eventually implement such predictive models, ensuring a safer and more secure urban area for all residents. This paper also advocates for the inclusion of more shared data (*e.g.* on a larger regional or national level) to support models in making more accurate predictions and filtering out excess noise (e.g. repeat callers, more specific categories, *etc.*). This is a direct result of implementing machine learning models, in addition to general data-analysis, to help law enforcement improve efficiency within their departments, which can be applied to managing resources, officer deployment, and more. While further research is needed to identify the effect of bias on such models, it is crucial for the future to leverage data-driven insights to improve the approach to community safety.

## Ⅷ. Ethical Considerations

The use of police call data and algorithmic predictions for proactive policing raises several ethical considerations. Firstly, there is a risk of reinforcing existing biases in the data. Historical police data may reflect systemic biases, such as racial profiling or socio-economic disparities, which can be exacerbated by a reliance on predictive models. This can lead to disproportionate targeting of certain communities, creating distrust between law enforcement and the public. Similarly, even though predictive models can be highly accurate, they still have the potential to produce harmful outcomes. False positives can result in unnecessary police interventions, causing undue stress and potential harm to individuals wrongly identified as threats. False negatives may lead to a lack of response to genuine threats, undermining public safety.

Privacy concerns are also a significant concern when working with predictive models, as they require the use of extensive and detailed data. The use of detailed personal data for predictive policing can infringe on individuals' privacy rights. There must be measures to ensure that data is anonymized and securely stored to prevent misuse or unauthorized access. The algorithms used for predictions should be transparent and open to discussion to ensure they are fair and unbiased. Law enforcement agencies must be accountable for the decisions made based on these predictions, with clear protocols for addressing errors and biases. Moreover, the ethical implications of surveillance must be considered. The widespread use of surveillance technologies, such as city cameras and social media monitoring, can infringe on civil liberties.

While machine learning and predictive models offer significant potential for enhancing public safety in the future, their implementation must be carefully managed to address ethical concerns. Balancing the benefits of proactive policing with the need to protect individual rights and prevent bias is essential for developing a fair and effective approach to crime prevention.

## IX. Acknowledgements

The author would like to acknowledge the help and support of Simeon Sayer at Harvard University for his mentorship on the approach and methodology throughout this research. The author also would also like to thank Anurag Srivastava for invaluable python tutoring and implementation tips during the project.